\documentclass[10pt,twocolumn,letterpaper]{article}

\usepackage{cvpr}
\usepackage{times}
\usepackage{epsfig}
\usepackage{graphicx}
\usepackage{amsmath}
\usepackage{amssymb}

\usepackage[pagebackref=true,breaklinks=true,letterpaper=true,colorlinks,bookmarks=false]{hyperref}

\cvprfinalcopy % *** Uncomment this line for the final submission

\def\cvprPaperID{2123} % *** Enter the CVPR Paper ID here

\begin{document}

%%%%%%%%% TITLE
\title{Feature Selective Networks for Object Detection}

\author{
Yao Zhai$^{1}$\thanks{This work was done during an internship at Microsoft Research Asia.}, Jingjing Fu$^2$, Yan Lu$^2$, and Houqiang Li$^1$\\
\\$^1$University of Science and Technology of China\\
$^2$Microsoft Research Asia\\
{\tt\small zy92918@mail.ustc.edu.cn, \{jifu, yanlu\}@microsoft.com, lihq@ustc.edu.cn}
}
\maketitle
%\thispagestyle{empty}

%%%%%%%%% ABSTRACT
\begin{abstract}
   Objects for detection usually have distinct characteristics in different sub-regions and different aspect ratios. However, in prevalent two-stage object detection methods, Region-of-Interest (RoI) features are extracted by RoI pooling with little emphasis on these translation-variant feature components. We present feature selective networks to reform the feature representations of RoIs by exploiting their disparities among sub-regions and aspect ratios. Our network produces the sub-region attention bank and aspect ratio attention bank for the whole image. The RoI-based sub-region attention map and aspect ratio attention map are selectively pooled from the banks, and then used to refine the original RoI features for RoI classification. Equipped with a light-weight detection subnetwork, our network gets a consistent boost in detection performance based on general ConvNet backbones (ResNet-101, GoogLeNet and VGG-16). Without bells and whistles, our detectors equipped with ResNet-101 achieve more than 3\% mAP improvement compared to counterparts on PASCAL VOC 2007, PASCAL VOC 2012 and MS COCO datasets.
\end{abstract}

%%%%%%%%% BODY TEXT
\section{Introduction}

\begin{figure}[t]
\includegraphics[width = .46\textwidth]{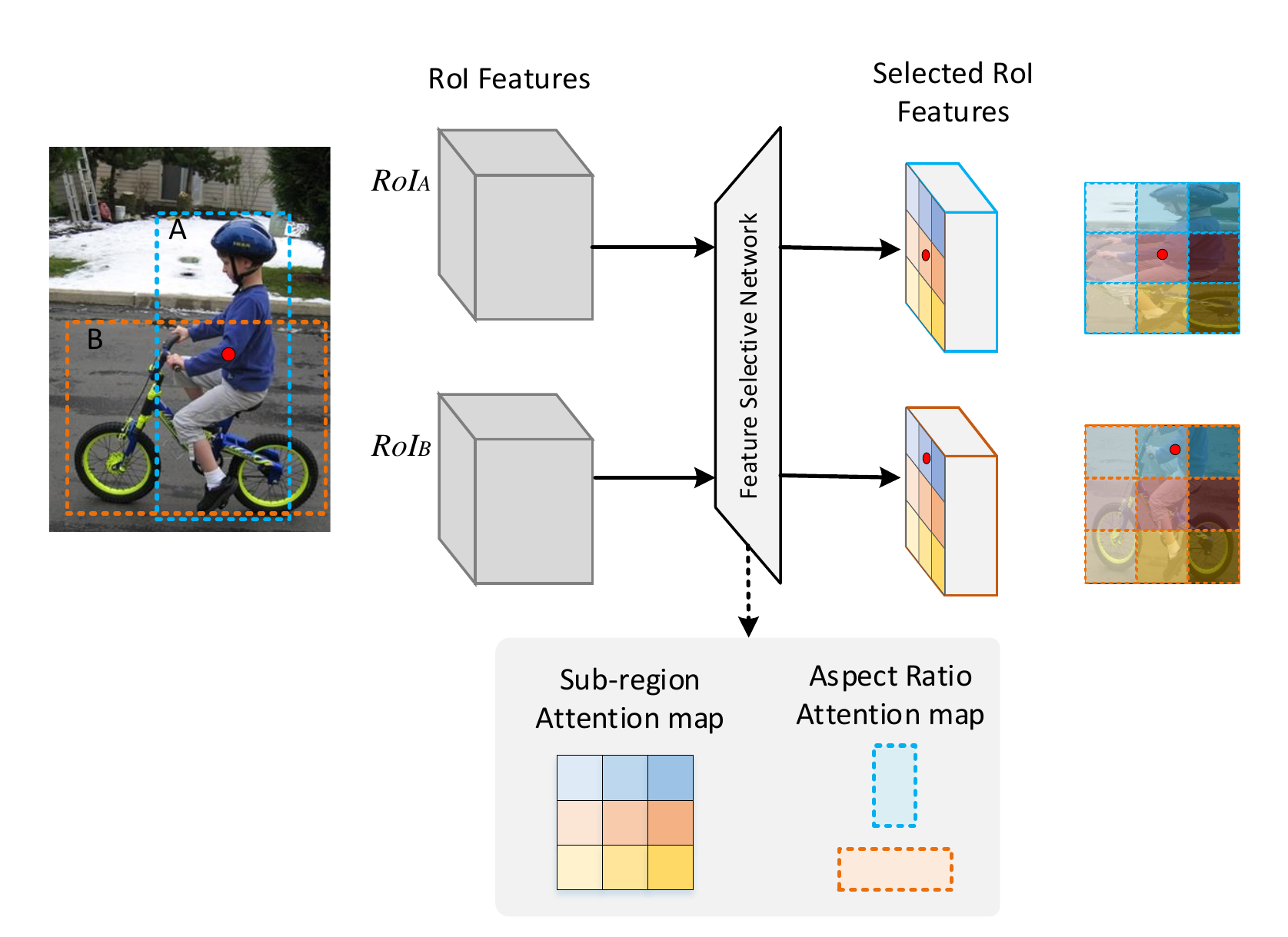}
\caption{We generate sub-region attention map and aspect ratio attention map for each RoI. Specifically, for the same spatial position (the red point) inside different sub-regions (center in $RoI_A$, top-center in $RoI_B$), the sub-region attention map gives different feature attentions. For the RoIs of different aspect ratios, the aspect ratio attention map also produces distinct feature attentions.}
\label{fig:1}
\end{figure}

Recent years have witnessed the rapid advancement of deep neural networks, which yields large performance improvements on image classification and detection tasks. ConvNets~\cite{alexnet,vgg,googlenet,resnet} designed for image classification have realized impressive representations of image features, outperforming traditional handcrafted features~\cite{hog,sift}. Object detectors adopting these deep ConvNets improve accuracy significantly on various detection benchmarks~\cite{voc,coco}. R-CNN~\cite{rcnn} firstly uses deep ConvNet to extract proposal region features by initializing parameters from a pre-trained ImageNet~\cite{imagenet} classification model. Fast R-CNN~\cite{fast-rcnn} develops a Region-of-Interest(RoI) pooling layer to extract RoI features from the convolutional feature maps of the entire image. After that, Faster R-CNN~\cite{faster-rcnn} introduces region proposal network (RPN) for generating accurate RoI proposals and sharing computation with detection subnetwork. Although one-stage object detection frameworks ~\cite{yolo,ssd,retina} have been proposed recently, most of state-of-the-art object detectors adopt two-stage framework composed of a proposal generator and a region classifier. Before RoI classification, RoI features are extracted with the following strategy: firstly generate whole image feature maps by ConvNets, and then pool the RoI features.

RoI features matter, as object detection relies on the classification and regression of RoI bounding boxes. Therefore, aside from the progress on deep ConvNets, research has focused on trying to generate powerful and informative RoI features to boost detection accuracy. To involve multi-scale features, HyperNet~\cite{hypernet} and FPN~\cite{fpn} produce RoI features by utilizing hierarchical feature maps from different depths of ConvNets. MR-CNN~\cite{mrcnn} and GBD-net~\cite{gbdnet} attempt to build a richer region-wise feature representation inside or around the RoI area. All these works adopt the classical RoI pooling layer, which divides RoIs into pooling bins and max-pools the corresponding spatial extent on convolutional feature map into fixed-length RoI features.

RoI features originating from convolutional feature map are high-dimensional ({\it e.g.}, 2048-d in ResNet-101). To model complex recognition patterns with high-dimensional features, heavy region classification networks are essential, leading to large-scale parameters and time consuming inference. Some might question whether RoI features aiming at region classification and regression actually need so many channels. In general, most of RoIs have already covered a considerable portion of objects thanks to accurate proposal generation~\cite{ss,locnet,edgebox,deepbox,deepproposal}. This means that it is possible to highlight the discriminative feature components out of deep features by RoI feature re-extraction.

Moreover, RoI features pooled by classical RoI pooling are \textit{translation-invariant}. Convolutional neural networks \textit{share weights spatially} over all positions on feature maps, forming the translation-invariant feature representations of the whole image. These translation-invariant features are insensitive to detection tasks that focus more on how to precisely localize objects. Notice that objects usually have distinct spatial characteristics in different sub-regions that draw on different feature representations. For example, boundary parts of an object may need more features describing edges and contours for localization, while the center parts favor texture features for classification. Additionally, objects from various categories or viewpoints may also keep varying aspect ratios of ground truth boxes. However, in classical RoI pooling, RoI features are extracted independently for different sub-regions and aspect ratios: features of different sub-regions are pooled on \textit{all channels of convolutional feature map}, with little emphasis on location-related components and aspect ratio preference.
%\iffalse
\begin{figure}[t]
\centering
\includegraphics[width = .5\textwidth]{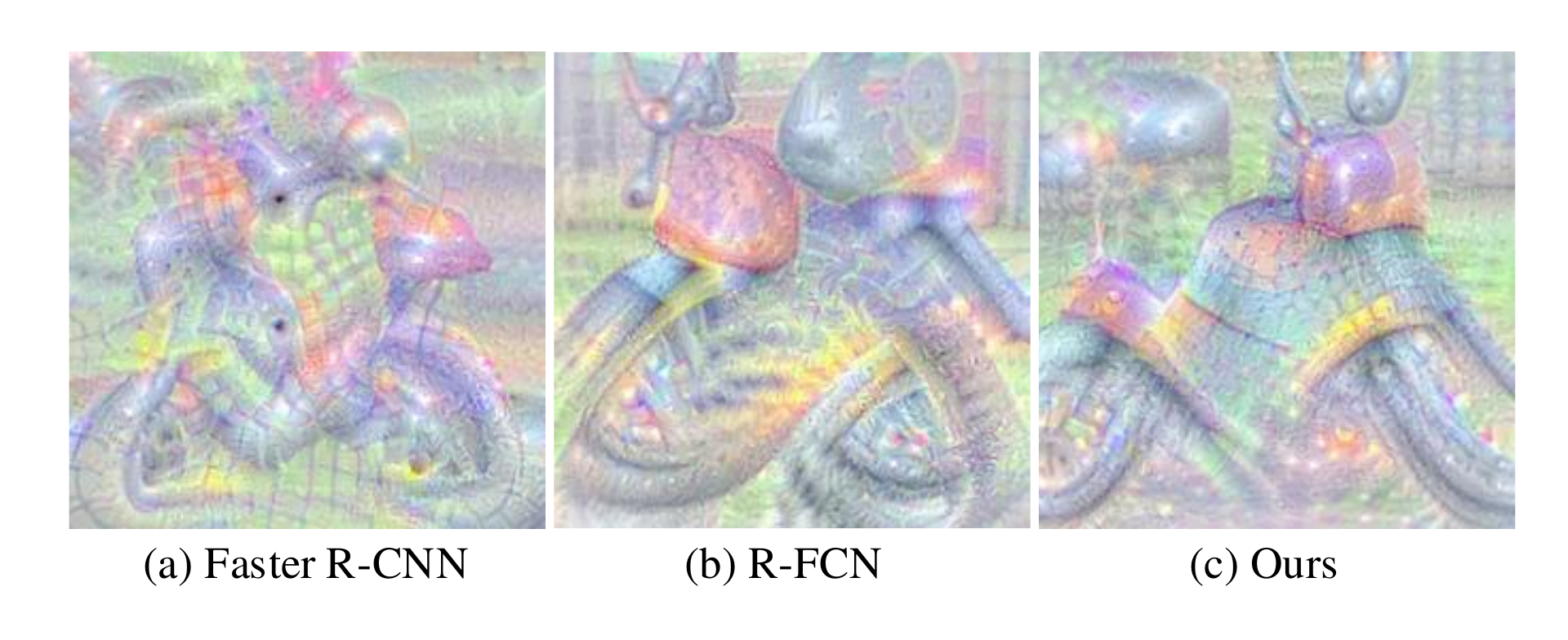}
\caption{Visualizations of object class ``motobike'' from the trained models on PASCAL VOC~\cite{voc} using DeepDraw~\cite{deepdraw}.}
\label{fig:2}
\end{figure}
%\fi

Based on the above observations, we propose \textit{feature selective networks}, which introduce dimension reduction and region-wise \textit{feature attention}. In our network, RoI features are extracted with respect to the sub-region variation and aspect ratio preference. Benefitting from intense dimension reduction, we replace the multiple high-capacity convolutional or fully connected ($\it{fc}$) layers in traditional region classifier by only one low-capacity $\it{fc}$ layer.

Figure~\ref{fig:1} shows a toy example of our feature selective network. Our network produces RoI feature representations with \textit{translation-variant} components based on the detailed sub-region and aspect ratio attention. The channel number $C_s$ ({\it e.g.}, 40) of RoI features in our network is much smaller than the channel number $C$ of the original convolutional feature map.

Figure~\ref{fig:2} shows the DeepDraw visualizations from the trained models of Faster R-CNN, R-FCN~\cite{rfcn} and our network. We can observe that our model well maintains structural characteristics and preserves the distinct spatial depiction of objects.

The effectiveness of our method is shown through experiments on PASCAL VOC and MS COCO~\cite{coco}. Our network gets a consistent boost in detection performance based on general ConvNet backbones (ResNet-101~\cite{resnet}, GoogLeNet~\cite{googlenet} and VGG-16~\cite{vgg}).

Our contributions can be summarized into three components:
\begin{itemize}
  \item Our network generates region-orientated attention maps and creates an informative \textit{translation-variant} representation of RoI features. Without elaborate enhancements, our detector equipped with ResNet-101 achieves more than 3\% mAP improvement compared to Faster R-CNN and R-FCN counterparts.
  \item A heavy detection subnetwork with multiple high-capacity $\it fc$ layers or convolutional layers is simplified to a single low-capacity $\it fc$ layer, which largely reduces parameter size and speeds up inference, especially for ResNet and GoogLeNet backbones.
  \item Our network is generic and gets a consistent boost based on different ConvNet backbones (ResNet-101, GoogLeNet and VGG-16).
\end{itemize}
%\iffalse
\begin{figure*}[t]
\centering
\includegraphics[width = 1.0\textwidth]{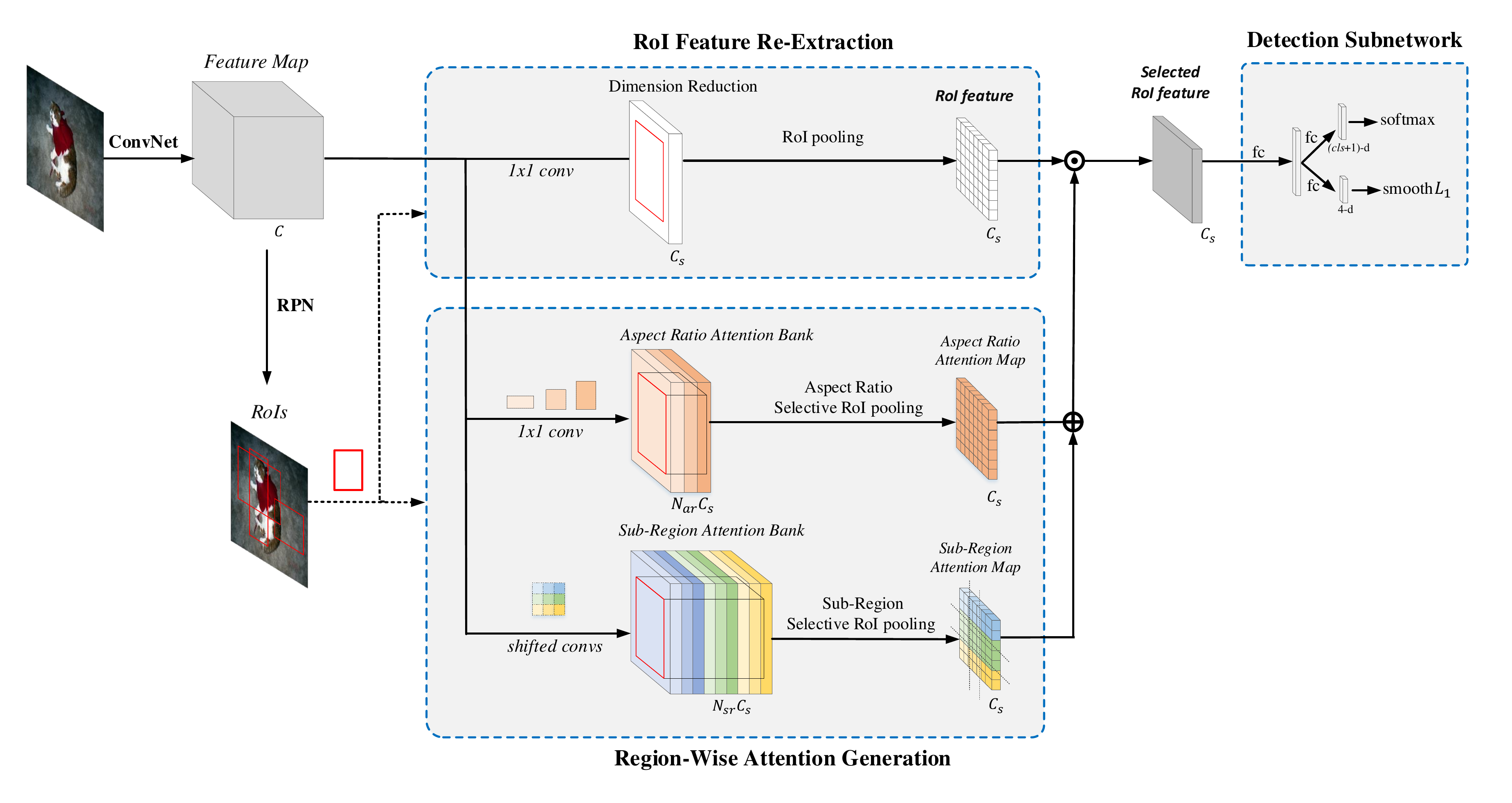}
   \caption{Architecture of our feature selective network. (1) Generate the convolutional feature map of the entire image by ConvNet and produce RoIs by RPN. (2) {\bf RoI Feature Re-Extraction}: Reduce the channel number of feature map from $C$ to $C_s$ ({\it e.g.}, $C=2048$, $C_s=40$) and perform RoI pooling to get the compacted RoI features. (3) {\bf Region-Wise Attention Generation}: Produce $N_{sr}C_s$-d \textit{sub-region attention bank} and $N_{ar}C_s$-d \textit{aspect ratio attention bank} according to RoI sub-region division ($N_{sr}=3\times3$) and aspect ratio division ($N_{ar}=3$), respectively. Then given an RoI, selectively pool the \textit{sub-region attention map} and \textit{aspect ratio attention map} referring its detailed sub-region and aspect ratio information (visualized as different colors in the figure, the pooling size is $7\times7$). (4) Generate the final selected RoI features by merging the attention maps with the compacted RoI features. (5) Feed the selected RoI features into a low-capacity detection subnetwork for RoI classification and regression.}
\label{fig:3}
\end{figure*}

%-------------------------------------------------------------------------
\section{Related Work}

\noindent {\bf Traditional Handcrafted Feature Extraction.} Deformable Parts Models (DPMs)~\cite{dpm,dpm2,dpm3} dominated object detection for years before CNN sprang up. DPM mainly adopts traditional handcrafted features upon all partitioned blocks of candidate boxes with latent SVM classifiers.

\noindent {\bf Convolutional Feature Extraction.} After the success of using deep neural networks for image classification~\cite{imagenet}, a research stream based on CNNs (OverFeat~\cite{overfeat}, R-CNN~\cite{rcnn}) shows significant improvements in detection accuracy. These methods use convolutional layers to extract features from each region proposal. To further speed up, SPP-Net~\cite{sppnet} and Fast R-CNN~\cite{fast-rcnn} firstly extract region-independent feature maps at the full-image level, and then pool region-wise features via spatial extents of proposals.

\noindent {\bf Region-Wise Feature Aggregation Networks.} To improve detection accuracy, several methods try to aggregate more effective region-wise features. MR-CNN~\cite{mrcnn} develops multiple region adaption modules to pool features from a candidate box's multi-regions. HyperNet~\cite{hypernet} integrates hierarchical feature maps together to generate hyper RoI features. Similarly, MS-CNN~\cite{mscnn} employs multi-scale layers for accurate proposal generation and classification. SDP~\cite{sdp} performs cascaded RoI pooling from different layers followed by corresponding RoI classifiers. ION~\cite{ion} exploits information both inside and outside RoIs along with four-directional IRNN represented contextual information. GBD-Net~\cite{gbdnet} addresses a gated bi-directional CNN to leverage features from multiple support regions.

\noindent {\bf Region-Based Fully Convolutional Networks.} After that, R-FCN~\cite{rfcn} and Deformable R-FCN~\cite{dcn} encode sub-region information in the detection framework by constructing a set of position-sensitive score maps. Instead of adopting a detection subnetwork, R-FCN computes the classification scores of sub-regions with a position-sensitive pooling layer and then averages them into the final RoI scores. R-FCN sufficiently performs multiple regression strategies for each sub-region, but fails to combine information of different sub-regions together. Figure~\ref{fig:2} shows that, R-FCN well maintains partial characteristics but has a hard time keeping global structural information of objects.

Inspired by R-FCN, our feature selective network exploits sub-region information by generating sub-region attention maps for selecting RoI features. Along with aspect ratio attention maps, these translation-variant attention maps empower RoI features to form operative feature representations for objects.

\section{Feature Selective Network}

\subsection{General Architecture}

Our goal is to extract effective RoI features from translation-invariant convolutional feature maps. To achieve this goal, we adopt the popular two-stage object detection framework that consists of a proposal generator and a region classifier. Figure~\ref{fig:3} shows an overview of our network architecture. Our network firstly forwards an input image through ConvNet and produces the convolutional feature map of the entire image. After that, the Region Proposal Network (RPN)~\cite{faster-rcnn} is grafted for generating candidate RoIs by anchor box classification and regression. Given the feature map and RoIs, we focus on the design of RoI feature extractor.

Our RoI feature extractor consists of RoI feature re-extraction and region-wise attention generation. Before region-wise attention generation, we adopt a $1\times1$ convolutional layer to reduce the channel number to $C_s$ and pool the compacted RoI features. Each RoI is assumed to be divided into $N_{sr}$ sub-regions for customized sub-region attention map extraction. According to sub-region division ($N_{sr}=3\times3$ demonstrated in Figure~\ref{fig:3}), we generate an $N_{sr}C_s$ -d sub-region attention bank for the entire image by a group of designed shifted convolutional layers. Likewise, we classify RoIs of different aspect ratios into $N_{ar}$ categories ($N_{ar}=3$ demonstrated in Figure~\ref{fig:3}) and then generate an $N_{ar}C_s$-d aspect ratio attention bank.

 Once provided the detailed information of an RoI, two designed selective RoI pooling layers max-pool the active attention maps from particular channel ranges in the sub-region attention bank and the aspect ratio attention bank, respectively. We aggregate the sub-region attention map $M_{sr}$ and aspect ratio attention map $M_{ar}$ together into the translation-variant attention map by element-wise addition. Finally, we merge the attention map with the compacted RoI features by element-wise product to get the selected RoI features.

 \begin{center}
$\hat{f}_i = f_i \cdot ({M_{sr}}_i+{M_{ar}}_i)$
\end{center}
\noindent
where $i = 1, ..., N$, and $N$ is the number of RoIs. For $i$-th RoI, $f_i$ is the compacted RoI features, and  $\hat{f}_i$ is the selected RoI features. ${M_{sr}}_i$ and ${M_{ar}}_i$ correspond to its sub-region and aspect ratio attention maps.

 A subsequent low-capacity detection subnetwork outputs the RoI classification score and class-agnostic box regression offsets. With efficient translation-variant RoI features, our network achieves state-of-the-art detection performance while maintaining a small parameter number and fast inference speed.

\subsection{Dimension Reduction}

Traditional object detectors employ ImageNet pre-trained classification backbones to extract region-independent features, followed by region-wise MLPs for RoI classification. The classical RoI pooling layer that acts as an RoI feature extractor generates fixed-length RoI features, whose channel number is huge without dimension reduction. Compared to the prevalent object detectors, our network re-extracts RoI features with a noticeably smaller channel number. We adopt a $1\times1$ convolutional layer to reduce the channel number of feature map from $C$ to $C_s$ and max-pool the RoI portion spatially into the original compacted RoI features. These compacted RoI features are expected to be optimized by later translation-variant attention maps.
%\iffalse
\begin{figure}[t]
\includegraphics[width = .5\textwidth]{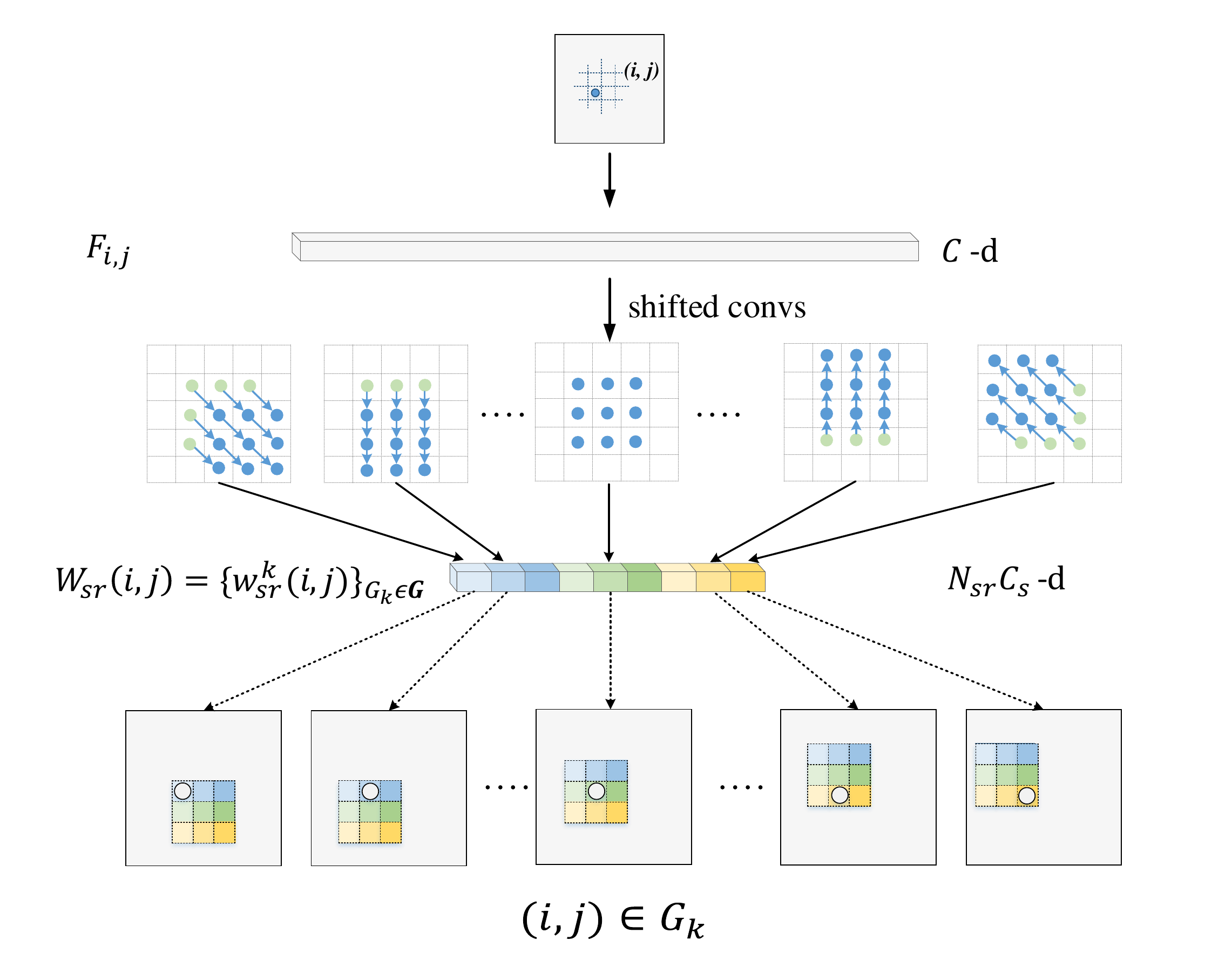}
\caption{Illustration of sub-region attention bank's generation.}
\label{fig:4}
\end{figure}
%\fi

\subsection{Attention Banks}

In order to develop translation-variant feature selection, we predict attention banks, which store all possible attention maps of spatial points when they are located in different sub-regions or RoIs of different aspect ratios.

\subsubsection{Sub-Region Attention Bank}

\indent Objects for detection usually exhibit distinct spatial characteristics in different sub-regions of an RoI (boundary, inner texture, surrounding context, {\it etc}.). However, features of these sub-regions are extracted position-independently by RoI pooling in most existing object detection methods. To address this issue, we generate a sub-region attention bank. In the sub-region attention bank, the attention vector at each position comprises of position-dependent components associated with its possible locations in RoI sub-regions.

The sub-region attention bank's generation process is illustrated in Figure~\ref{fig:4}. An RoI denoted as $\it G$ is divided into $N_{sr} = 3\times3$ sub-regions, represented by ${\{G_k\}}_{1\leq k\leq N_{sr}}$. The feature vector at the $(i,j)$-th spatial position of convolutional feature map is denoted as $F_{i,j}$ with $C$ channels. The active sub-region attention at $(i,j)$ are obtained by,
\begin{center}
$w_{sr}^k(i,j) = {\bf \Phi}_k (F_{i,j} )$
\end{center}

\noindent $w_{sr}^k(i,j)$ is the active attention vector with $C_s$ channels assuming that the position $(i,j)$ is located in the spatial extent of $k$-th sub-region $G_k$ of an RoI. ${\bf \Phi}_k$ represents the feature attention extractor for $G_k$, which is implemented in the form of the $k$-th shifted convolutional layer. The parameters of ${\bf \Phi}_k$ are learned by backpropagation. The sub-region attention bank $W_{sr}$ at the position $(i,j)$ is a catenation of active feature attentions $\{w_{sr}^k(i,j)\}_{1\leq k\leq N_{sr}}$. Hence, the total size of sub-region attention bank is $H\times{W}\times{N_{sr}C_s}$. $H$ and $W$ are the spatial sizes of the original convolutional feature map. The total channel number of sub-region attention bank is $N_{sr}C_s$.

On the original convolutional feature map, every spatial position shares the same representations encoded in $C$ channels. Instead, in the sub-region attention bank, every spatial position has a set of customized attention values containing sub-region details. For the spatial position $(i,j)$, channels ranging from $(k-1)C_s+1$ to $kC_s$ indicate the specific feature attentions when $(i,j)$ is located in the spatial extent of the $k$-th sub-region of an RoI.

For a concrete example, when setting the selective channel number to 40, we generate an attention bank with a total of 360 channels. Each set of 40 channels for a spatial position serves as the active attention values when the position is inside an RoI's sub-regions of \textit{top-left}, \textit{top-center}, \textit{top-right}, ..., \textit{bottom-center} and \textit{bottom-right}, respectively. Intuitively, in the following feature-selective RoI pooling layer, each feature map point inside an RoI will correspond to a group of 40 channels in the sub-region attention bank, according to the detailed relative position.

\vspace{6pt}
\noindent {\bf Shifted Convolution.} We design a group of shifted convolutional layers on the convolutional feature map of the entire image to produce the sub-region aware attention bank. The shifted convolutions are special cases of deformable convolutions in~\cite{dcn}. Different from deformable convolution, shifted convolution keeps the same 2D offsets for different spatial positions on feature map. As shown in Figure~\ref{fig:4}, the 2D offsets of shifted convolutional layers are fixed to $(1,1), (1,0), (1,-1), ..., (-1, -1)$, respectively. The shift directions are aligned with the directions that the corresponding sub-regions towards the RoI center. The convolution kernel is fixed to $3\times3$.
%\iffalse
\begin{figure}[t]
\includegraphics[width = .5\textwidth]{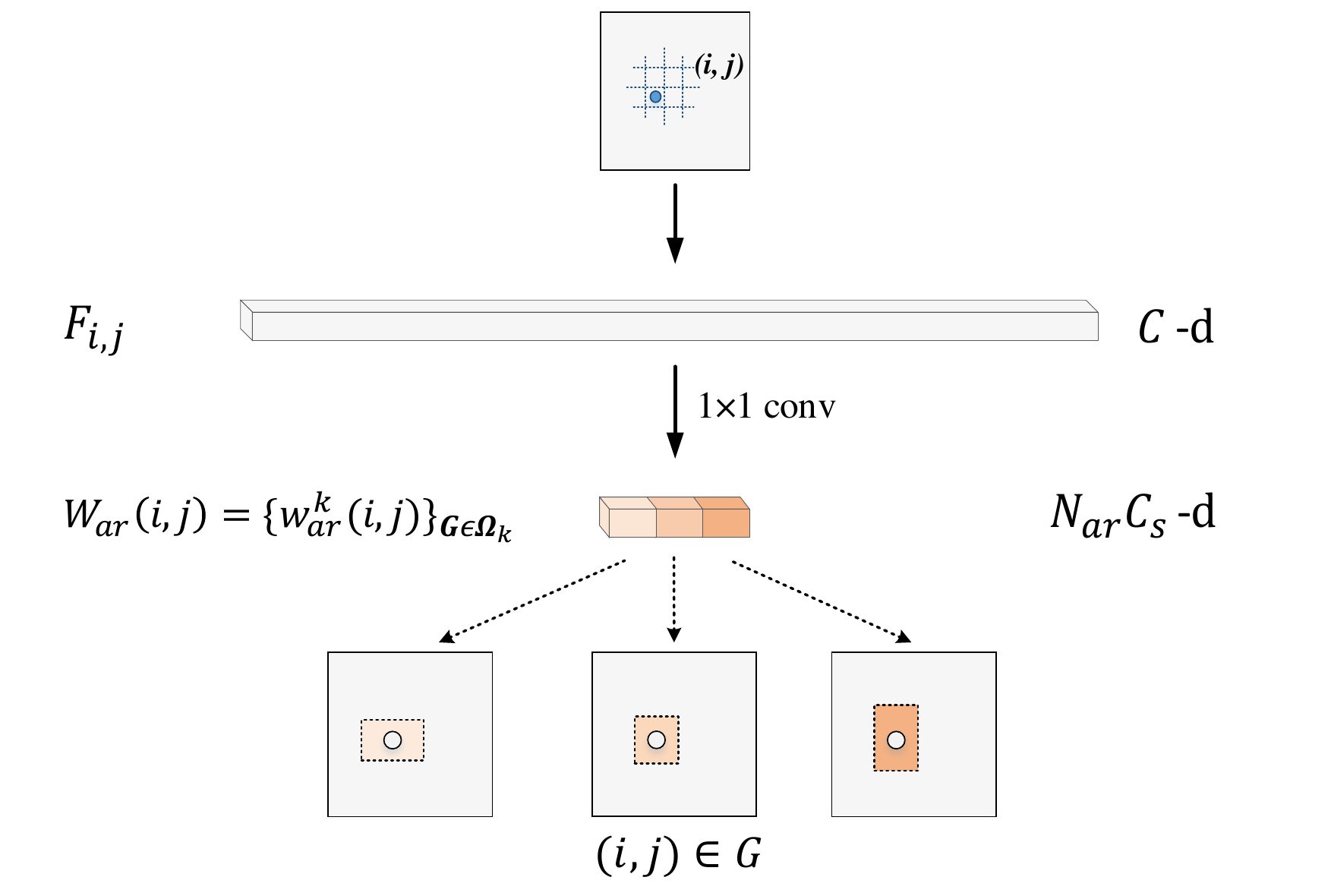}
\caption{Illustration of aspect ratio attention bank's generation.}
\label{fig:5}
\end{figure}
%\fi

\subsubsection{Aspect Ratio Attention Bank}

\indent Sub-region attention bank explores the potentials of position information inside RoIs. Apart from sub-regions, our network also takes into account aspect ratio information. In practice objects of different categories usually carry different aspect ratios. Besides objects difference, aspect ratio information may also reflect the viewpoint or pose of an object. For instance, there are large differences in aspect ratio between pedestrians and sitting people. Therefore, aspect ratio information should also be considered in RoI feature extraction. However, the classical RoI pooling layer generates a fixed spatial size ({\it e.g.}, $7\times7$) feature representation for all RoIs, thereby ignoring the aspect ratio difference between them.

To remedy this issue, we produce the aspect ratio attention bank to utilize the aspect ratio information. In parallel with sub-region attention bank, a $1\times1$ convolutional layer is placed on the convolutional feature map to get the aspect ratio aware components of each spatial position. The generation process is shown in Figure~\ref{fig:5}. We group RoIs of different aspect ratios into $N_{ar}$ categories ($N_{ar}$ is set to 3 in Figure~\ref{fig:5}: ratio $<$ 0.75, ratio $>$ 1.3 and others). $w_{ar}^k$ with $C_s$ channels is derived from $F_{i,j}$, inferring the active feature attention if the position $(i,j)$ is located in the RoI categorized to the $k$-th RoI set ${\bf \Omega}_{k}$. The aspect ratio attention bank $W_{ar}$ at $(i,j)$ is a catenation of active feature attentions $\{w_{ar}^k(i,j)\}_{1\leq k\leq N_{ar}}$.

\subsection{Attention Maps}

In our feature selective network, attention banks are produced to reveal the particular and distinct characteristics of different sub-regions and aspect ratios. Once candidate RoIs are provided by RPN, we perform \textit{selective RoI pooling} layers to dedicate the sub-region and aspect ratio details to translation-variant attention maps. Our selective RoI pooling layer leverages max-pooling to map an RoI's spatial extent on attention banks into a fixed-length attention vector of $h\times{w}\times{C_s}$ ({\it e.g.}, $7\times7\times40$). $h$ and $w$ are the pooling sizes, and $C_s$ is the selective channel number.

\vspace{6pt}
\noindent{\bf Selective RoI Pooling.} Specifically, in an RoI window, $(m,n)$-th pooling bin has a correspondence with indexes $k_{sr}$ ($1\leq k\leq N_{sr}$) and $k_{ar}$ ($1\leq k\leq N_{ar}$) for sub-region attention bank and aspect ratio attention bank, respectively. For the sub-region attention bank, index $k_{sr}$ denotes that most area of the pooling bin belong to the spatial rectangular $G_{k_{sr}}$ of the $k_{sr}$-th sub-region. Generally, \small{${\bf bin}(1,1)\in G_1,...,{\bf bin}(m,n)\in G_{k_{sr}},..., {\bf bin}(h,w)\in G_{N_{sr}}$}. \normalsize{For the aspect ratio bank, index $k_{ar}$ denotes that the RoI is a member of $k_{ar}$-th aspect ratio set ${\bf \Omega}_{k_{ar}}$. Consequently, $h\times{w}$ pooling bins share the same $k_{ar}$. Given $k_{sr}$ and $k_{ar}$, we max-pool the attention values from particular channel lists of the spatial positions inside each pooling bin.}
\begin{center}
$M_{sr}(m,n,c) = \max\limits_{(i,j)\in {\bf bin}(m,n)}W_{sr}(i,j,c+(k_{sr}-1)C_s)$
$M_{ar}(m,n,c) = \max\limits_{(i,j)\in {\bf bin}(m,n)}W_{ar}(i,j,c+(k_{ar}-1)C_s)$

\end{center}
where $M_{sr}(m,n,c)$ and $M_{ar}(m,n,c)$ respectively stand for the $c$-th channel of the $(m,n)$-th pooling bin for sub-region attention map and aspect ratio attention map ($1\leq m\leq h, 1\leq n\leq w, 1\leq c\leq C_s$). According to the indexes $k_{sr}$ and $k_{ar}$, the particular channel lists range from $1+(k_{sr}-1)C_s$ to $k_{sr}C_s$ in the sub-region attention bank $W_{sr}$ and $1+(k_{ar}-1)C_s$ to $k_{ar}C_s$ in the aspect ratio attention bank $W_{ar}$, separately.

Hence, each pooling bin of the RoI selects a distinctive $C_s$-d sub-region dependent attention map. RoIs of different aspect ratios also select a particular $C_s$-d aspect ratio aware attention map. So far, we produce two attention maps with fixed-length of $h\times w\times C_s$ as the representations of sub-region and aspect ratio disparities. We merge the two maps to get the translation-variant attention map by element-wise addition. After weighting the original dimension-reduced RoI features, we feed the selected RoI features into a lightweight detection subnetwork to get the classification and regression outputs.

During backpropagation, the selective RoI pooling layer follows the chain rule and propagates the gradients from attention maps to the specific position in attention banks, according to the aforementioned correspondence.

\subsection{Detection Subnetwork}

In general, the detection subnetworks of two-stage object detectors are high-capacity or deep. The high-capacity ones are originated from AlexNet and VGG, which have two 4096-d $\it fc$ layers designed for image classification. For detectors adopting superior fully convolutional classification backbones ({\it e.g.}, ResNet, GoogLeNet), a deep and convolutional detection subnetwork is necessary~\cite{noc}.

Enabled by concise and informative RoI feature re-extraction, we simplify our detection subnetwork to a single low-capacity ({\it e.g.}, 500-d) $\it fc$ layer, followed by a ($cls$+1)-d $\it fc$ layer to get RoI classification scores and a 4-d $\it fc$ layer to output class-agnostic bounding box regression offsets. Our detection subnetwork has a much smaller parameter size than the classical RoI classifier. At the same time, this simple design overcomes the drawback of costing inference time when the detection subnetwork has a deep network architecture.

\section{Experiments}

Our feature selective network is a generic and effective method of extracting RoI features for two-stage object detectors. In the subsequent experiments, we attach our network to commonly used ConvNet backbones: ResNet-101, GoogLeNet and VGG-16.

\subsection{Experimental Setup}

\noindent {\bf ConvNet Backbones.} Original ResNet-101 and GoogLeNet, designed for image classification, are fully convolutional and have a stride of 32 pixels on the last convolutional layers. We follow the modification in NoC~\cite{noc}: reducing the stride from 32 to 16 by changing the last stride operation from 2 to 1 and applying the ``hole algorithm''~\cite{fcn,hole}(\emph{``Algorithme \`{a} trous''}~\cite{atrous}). We then attach our network to the last convolutional layers of ConvNet backbones.

\noindent {\bf Proposal Subnetwork.} For region proposal generation, we use RPN to generate 300 proposals per image, and then perform all ablation experiments on fixed proposals obtained from the RPN of three ConvNets, respectively.

\noindent {\bf Detection Subnetwork.} We feed the RoI feature into one added $\it fc$ layer for feature reorganization, followed by a ($cls$+1)-d $\it fc$ layer and a 4-d $\it fc$ layer to get outputs. The dimension of added $\it fc$ layer is 500-d, or changed to 100-d if RoI features only have one channel. We use softmax loss and smooth $L_1$ loss defined in Fast R-CNN~\cite{fast-rcnn} for backpropagation~\cite{bp}.

\noindent {\bf Training Details.} We start from ImageNet pre-trained models of each ConvNet, and label an RoI as foreground when its Intersection over Union (IoU) overlap with a ground truth box is at least 0.5. Other RoIs are labeled as background. During the training process, we construct each minibatch from two images with 256 RoIs and allocate 25\% of them to the foreground. Image or its horizontal flip are resized to a single scale that its shorter side has 600 pixels.

\noindent {\bf Testing Details.} We test all images on single scale with shorter side 600 pixels, and adopt non-maximum suppression (NMS) with an IoU threshold of 0.3, then evaluate the results on the test benchmarks: PASCAL VOC 2007 test, VOC 2012 test, and MS COCO val2015 sets.

\subsection{Compared to baselines}

\begin{table}[t]
\begin{center}
\begin{tabular}{l|c|c|c}
\hline
method & sub-region? & aspect ratio? & mAP(\%) \\
\hline
Faster R-CNN &  &  & 78.8 \\
R-FCN~\cite{rfcn} &  &  & 79.5\\
\hline
Ours(a) & $\checkmark$ &  & 82.2 \\
Ours(b) &  & $\checkmark$ & 81.0 \\
Ours(c) & $\checkmark$ & $\checkmark$ & {\bf 82.9} \\
\hline
\end{tabular}
\end{center}
\caption{Detection results on VOC 2007 test set using ResNet-101. The training set is VOC 07+12 trainval.}
\label{lab:2}
\end{table}

\noindent {\bf ResNet-101.} We implement our algorithm with ResNet-101 on three settings. We divide RoIs into $3\times3$ sub-regions and three aspect ratio groups (ratio $<$ 0.75, ratio $>$ 1.3 and others). We generate attention banks by convolutional layers after res5c, and then select 40 channels for each pooling bin during feature-selective RoI pooling layers. To facilitate comparison, Ours(a) only uses sub-region attention map, Ours(b) only applies aspect ratio attention map, and Ours(c) combines sub-region attention map and aspect ratio attention map together.
Table~\ref{lab:2} shows the VOC 2007 test results of the reimplemented Faster R-CNN, R-FCN and our network using ResNet-101. R-FCN~\cite{rfcn} with OHEM achieves 79.5\% mAP. By contrast, Ours(a) with only sub-region information improves the baselines by 2.7 points. With the assistance of aspect ratio information, we further gain 82.9\% mAP, outperforming the baselines over 3 points. These results verify the effectiveness of the region-wise attention maps generated by our feature selective networks.

\begin{table}[t]
\begin{center}
\begin{tabular}{l|c|c|c|c}
\hline
method & Network & sub-   & aspect & mAP \\
       &         & region? & ratio?  & (\%) \\
\hline
Faster R-CNN & GoogLeNet & & & 74.8 \\
\hline
Ours(a) & GoogLeNet & $\checkmark$ & & 76.4 \\
Ours(b) & GoogLeNet &  & $\checkmark$ & 75.6 \\
Ours(c) & GoogLeNet & $\checkmark$ & $\checkmark$ & {\bf 76.8} \\
\hline \hline
Faster R-CNN & VGG-16 & & & 73.2 \\
\hline
Ours(a) & VGG-16 & $\checkmark$ &  & 73.9\\
Ours(b) & VGG-16 &  & $\checkmark$ & 73.6\\
Ours(c) & VGG-16 & $\checkmark $ & $\checkmark$ & {\bf 74.3}\\
\hline
\end{tabular}
\end{center}
\caption{Detection results on VOC 2007 test set using GoogLeNet and VGG-16. The training set is VOC 07+12 trainval.}
\label{lab:3}
\end{table}

\noindent {\bf GoogLeNet and VGG-16.} We conduct more experiments on GoogLeNet and VGG-16 with the same settings as in ResNet-101. We attach our network to the last convolutional layers: inception5b and conv5\_3, then extract $7\times7\times40$-d features for each RoI. To reimplement Faster R-CNN with GoogLeNet, we insert an RoI pooling layer after inception4e and employ later convolutional layers as detection subnetwork. Table~\ref{lab:3} shows the VOC 2007 test results on GoogLeNet and VGG-16. Faster R-CNN with GoogLeNet achieves 74.8\% mAP. In comparison, our network achieves 76.8\%, 2.0 points higher than the baseline. For the VGG-16 network, our method also yields a slight improvement. Notice that, Faster R-CNN inherits the two high-capacity 4096-d $\it fc$ layers from VGG-16, thus has a much larger model size than our detector. The experiments on GoogLeNet and VGG-16 indicate that the feature selective module is generic and robust to prevalent ConvNets. The ranges of performance improvements reflect the feature representation potential of each ConvNet, and our feature selective modules exploit the potentials to some extent.

\subsection{VOC 2012 Results}
\begin{table}[t]
\begin{center}
\begin{tabular}{l|c|c|c}
\hline
method &  sub-region? & aspect ratio? & mAP(\%) \\
\hline
Faster R-CNN & & & 73.8 \\
R-FCN & & & 77.6 \\
\hline
Ours(a) & $\checkmark$ & & 79.5 \\
Ours(b) & & $\checkmark$ & 77.9  \\
Ours(c) & $\checkmark$ & $\checkmark$ & {\bf 80.5}  \\
\hline
\end{tabular}
\end{center}
\caption{Detection results on VOC 2012 test set using ResNet-101. The training set is VOC 07+12 trainval+07 test.}
\label{lab:4}
\end{table}

\indent Table~\ref{lab:4} shows the results of VOC 2012 test using ResNet-101. We evaluate the performances with the same settings as in VOC 2007 test, dividing RoIs to $3\times3$ sub-regions, 3 aspect ratio ranges and selecting 40 channels for RoI features. R-FCN~\cite{rfcn} with multiscale training and OHEM~\cite{ohem} obtains 77.6\% mAP. Compared to that, our method achieves 80.5\% mAP.

\subsection{MS COCO Results}

\begin{table}[t]
\begin{center}
\begin{tabular}{l|c|c|c|c|c}
\hline
method & $\mathit{AP}$@0.5 & $\mathit{AP}$ & $\mathit{AP}_s$ & $\mathit{AP}_m$ & $\mathit{AP}_l$\\
\hline
Faster R-CNN & 48.4 & 27.2 & 6.6 & 28.6 & 45.0 \\
\hline
R-FCN & 48.9 & 27.6 & 8.9 & 30.5 & 42.0 \\
\hline
Ours & {\bf 54.0} & {\bf 33.6} & {\bf 17.8} & {\bf 35.4} & {\bf 46.5} \\
\hline
\end{tabular}
\end{center}
\caption{Detection results on the MS COCO val 2015 set using ResNet-101. The training set is COCO train 2015.}
\label{lab:5}
\end{table}

\indent We perform experiments on the challenging MS COCO dataset that has 80k train images and 40k val images. Since COCO has more complicated object categories, we make a modification that selectively pool 80 channels for each RoI. For the detection subnetwork, we repurpose the dimension of the $\it fc$ layer into 800-d. Table~\ref{lab:5} shows that our network achieves a better performance (54.0\% / 33.6\%) compared to Faster R-CNN~\cite{resnet} (48.4\% / 27.2\%) and R-FCN~\cite{rfcn} (48.9\% / 27.6\%) baselines. Our feature selective network far surpasses other methods on small size object detection.

\subsection{Parameter Number and Inference Speed}

\begin{table}[t]
\begin{center}
\begin{tabular}{l|l|c|c}
\hline
method & \# params & inference speed & mAP \\
       & &(\small{sec/img}) & (\%) \\
\hline
Faster R-CNN  & {\bf 40.7M} & 0.31 & 78.7\\
\hline
R-FCN  & 44.0M & {\bf 0.12}& 79.5\\
\hline
Ours (1x1 conv) &  42.6M & 0.19 & {\bf 82.3} \\
\hline

\end{tabular}
\end{center}
\caption{Comparisons in parameter number, inference speed and detection accuracy on VOC 2007 test set using ResNet-101. The training set is VOC 07+12 trainval.}
\label{lab:8}
\end{table}

Table~\ref{lab:8} shows the overall comparisons in parameter number, inference speed and the performance. Here our network adopts $1\times1$ convolution as the trade-off between performance and model complexity when generating sub-region attention map. The selective channel number is set to 40. We record the inference speeds on a Titan X GPU. Our network gains a better performance while maintaining a smaller parameter number compared to R-FCN. On the other hand, the deep detection subnetwork leads to significant per-RoI computation cost for Faster R-CNN. With the simplified detection subnetwork, our network has a higher inference speed than Faster R-CNN.

\section{Ablation studies}

\subsection{Dimension Reduction}

Before region attention map generation, our feature selective network adopts a $1\times1$ convolutional layer to reduce the channel number of convolutional feature map and RoI-pools the original compacted RoI features. Here we perform the ablation experiments on the role of dimension reduction. We reimplement Faster R-CNN with ResNet-101 using the two settings in~\cite{noc}: one uses res4b22's output as RoI feature and employ conv5 block as detection network, the other one adds two 4096-d fc layer as RoI classifier.

\begin{table}[t]
\begin{center}
\begin{tabular}{l|c|c|c}
\hline
feature & channel & RoI classifier & mAP(\%) \\
\hline
\small{res4b22} & 1024 & \small{res5a, 5b, 5c, fc21} & 78.7 \\
\hline
\small{res5c} & 2048 & \small{fc4096, fc4096, fc21} & 78.8 \\
\hline
\small{res5c} & 2048 & \small{fc500, fc21} & 78.4 \\
\hline
\small{res5c} & 40 & \small{fc500, fc21} & {\bf 79.5} \\
\hline

\end{tabular}
\end{center}
\caption{Detection results on VOC 2007 test set using Faster R-CNN with ResNet-101. The training set is VOC 07+12 trainval.}
\label{lab:1}
\end{table}

Table~\ref{lab:1} shows that, with a simplified detection subnetwork, dimension reduction achieves 79.5\% mAP on VOC 2007 test set, which is a 0.7\% performance improvement than the baseline. This indicates that a lightweight RoI classifier with compacted RoI features may receive a better detection result.

\subsection{Selective Channel Number}

Our feature selective network ensures a promising detection accuracy with a small channel number of RoI features. Here we investigate on the role of selective channel number $C_s$. We follow the same settings as before but change the selective channel number from 100 to 1 and report the results. Table~\ref{lab:6} shows that, enabled by sub-region and aspect ratio attention maps, our networks have similar results when the selective channel number $C_s$ varies from 20 to 100. 20 channels seem to be enough for RoI feature representation. Surprisingly, even reduced to one channel, the re-extracted RoI feature with a fixed-length of 49 still performs well.
\begin{table}[t]
\begin{center}
\begin{tabular}{l|c|c|c|c|c}
\hline
channel number $C_s$ &100&40&20&5&1 \\
\hline
mAP(\%) &82.9&82.9&82.5&80.5&{\bf 79.4} \\
\hline
\end{tabular}
\end{center}
\caption{Detection results on VOC 2007 test set using ResNet-101 with different selective channel numbers. The training set is VOC 07+12 trainval.}
\label{lab:6}
\end{table}

\subsection{Shifted Convolution}

Our feature selective network adopts shifted convolution operation when generating sub-region attention bank. The shifted convolution enables different feature extraction ways when a feature map point is inside different sub-regions. Here we investigate on the influence of shifted convolution operation. We adopt $1\times1$ and $3\times3$ standard convolution when generating sub-region attention bank. The results in table~\ref{lab:7} shows that, $3\times3$ convolution yields 0.3\% mAP gain than $1\times1$ benefitting from the increase of parameters. When equipped with shifted convolution, our network further achieves a 0.3\% mAP gain. The shift direction is towards the assumed RoI's center. If we change the shift direction to the opposite or random, evolving more features outside the RoI, the shifted convolution could hardly achieve better results than standard convolution. These results indicate that when extracting sub-region feature attentions for RoI, the feature information inside an object may play a more important role.
\begin{table}[t]
\begin{center}
\begin{tabular}{c|c|c|c}
\hline
conv kernel & shifted ? & shift direction & mAP(\%) \\
\hline
$1\times1$ &  & - & 82.3\\
$3\times3$ &  & - & 82.6 \\
\hline
$3\times3$ & $\checkmark$ & center & {\bf 82.9}\\
$3\times3$ & $\checkmark$ & outside & 82.5\\
$3\times3$ & $\checkmark$ & random & 82.5\\
\hline
\end{tabular}
\end{center}
\caption{Detection results on VOC 2007 test set using ResNet-101 with different convolution settings. The training set is VOC 07+12 trainval.}
\label{lab:7}
\end{table}

\section{Conclusion}

We propose feature selective networks to distill effective RoI features from convolutional feature maps. By generating attention banks, we exploit the translation-variant potential of RoI feature representations. Based on the detailed sub-region and aspect ratio information, distinctive attention maps are selected for each RoI and used to refine the original compacted RoI features. With a surprisingly small channel number 1-d for RoI features, our feature selective network ensures a state-of-the-art detection accuracy. With a proper selective channel number, our networks further achieve general improvements equipped with the prevalent ConvNet backbones (ResNet-101, GoogLeNet and VGG-16). Our method offers a general and efficient module to dedicate RoI preference to object detection networks.

{\small
\bibliographystyle{ieee}
\bibliography{ref}
}

\end{document}